\pdfoutput=1

\documentclass[11pt]{article}

\usepackage{ACL2023}

\usepackage{times}
\usepackage{latexsym}

\usepackage[T1]{fontenc}

\usepackage[utf8]{inputenc}

\usepackage{microtype}

\usepackage{inconsolata}

\usepackage{placeins} 
\usepackage{multirow}
\usepackage{graphicx}
\usepackage{amsfonts} 
\usepackage{amsmath} 
\usepackage{xcolor} 
\definecolor{LightGray}{gray}{0.7}

\usepackage{hyphenat}
\newcommand{\R}[1]{\overline{\textsc{r}_{#1}}}
\newcommand{\RS}{\R{\Sigma}}

\mathchardef\ordinarycolon\mathcode`\:
\mathcode`\:=\string"8000
\begingroup \catcode`\:=\active
  \gdef:{\mathrel{\mathop\ordinarycolon}}
\endgroup

\title{\vspace{-1.0cm}Detecting Idiomatic Multiword Expressions in Clinical Terminology \\ using Definition-Based Representation Learning}

\author{François Remy \\
  Ghent University - Imec \\
  \texttt{francois.remy@ugent.be} \\\And
  Alfiya Khabibullina \\
  University of Malaga \\
  \texttt{0611289993@uma.es} \\\And
  Thomas Demeester \\
  Ghent University - Imec \\
  \texttt{thomas.demeester@ugent.be} \\}

\begin{document}
{
    \maketitle
    \vspace*{-1cm}
}
\begin{abstract}
This paper shines a light on the potential of definition-based semantic models for detecting idiomatic and semi-idiomatic multiword expressions (MWEs) in clinical terminology. Our study focuses on biomedical entities defined in the UMLS ontology and aims to help prioritize the translation efforts of these entities. In particular, we develop an effective tool for scoring the idiomaticity of biomedical MWEs based on the degree of similarity between the semantic representations of those MWEs and a weighted average of the representation of their constituents. We achieve this using a biomedical language model trained to produce similar representations for entity names and their definitions, called BioLORD. The importance of this definition-based approach is highlighted by comparing the BioLORD model to two other state-of-the-art biomedical language models based on Transformer: SapBERT and CODER. Our results show that the BioLORD model has a strong ability to identify idiomatic MWEs, not replicated in other models. Our corpus-free idiomaticity estimation helps ontology translators to focus on more challenging MWEs.
\vspace{-0.01cm}
\end{abstract}

\section{Introduction}

\begin{figure*}[t!]
    \centering
    \includegraphics[width=1.0\textwidth]{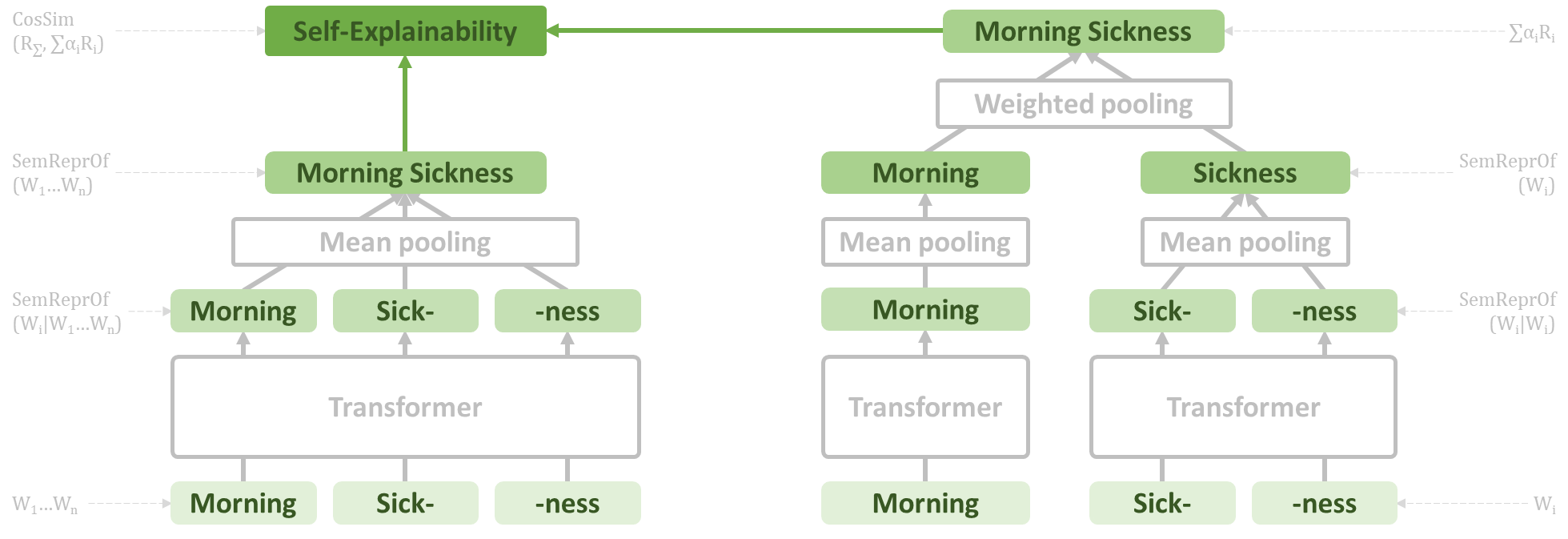} 
    \caption{
        \mbox{\makebox[0.91\linewidth][s]{\hspace{-0.05cm}In this paper, we use a cosine similarity metric to compare the representation of a MWE with the weighted}} average of the representations of its two constituents, after embedding each of these with the same semantic model which is based on a Transformer pipeline. Any difference in representation between these must come 
        from interactions between constituents within the Transformer when these constituents are combined in the MWE.
    }
    \label{fig:model}
\end{figure*}

Translation in the biomedical domain remains particularly challenging due to the large number of specific and ad-hoc usage of terminology \citep{neves-etal-2018-findings,neves-etal-2022-findings}. Some medical ontologies such as UMLS \citep{UMLS} contain more than 4 million entities. Out of these, only a fraction has already been labelled in languages other than English. While large efforts to translate some medical ontologies such as SnomedCT \cite{SnomedCT} can be noted, few if any of these efforts have yet to yield full coverage of the ontology in their target language \citep{Macary2020,Auwers2020}. 

Popularity is of course one factor motivating the prioritization of the expert translation of some entity names over others, as translating popular entities makes the ontology usable to a large number of practitioners at a lower cost. But, with the rise of automatic translation tools, another factor worth considering in the prioritization is the translation difficulty of the entities being passed on to medical translation experts. Their efforts should indeed better be directed to cases where automatic translation does not provide good results.

In this context, idiomaticity has a key role to play. Indeed, the automatic translation of idiomatic\footnote{MWEs are referred to as idiomatic if their meaning cannot be deduced from the interpretation of their constituents, in line with the definition of "Multiword Terms" presented by \citet{Ramisch2010}; examples in the biomedical domain include "Gray Matter" or "Morning Sickness".} MWEs poses a significant challenge, as juxtaposing the translation of each individual constituent often results in a loss of meaning that can, in some cases, be catastrophic. This difficulty has been noted by prominent researchers such as \citet{koehn-knowles-2017-six} and \citet{Evjen2018}. As a result, identifying such idiomatic MWEs would therefore immensely benefit the prioritization of translation efforts of medical ontologies. 


While many strategies for identifying MWEs have been presented in the past \citep{Ramisch2010,Kafando2021,zeng-bhat-2021-idiomatic}, we found that applying them to the medical domain (and especially its clinical counterpart) was challenging due to the extreme corpus size that would be required to produce statistically significant results for the long tail of medical entities.

In this paper, we investigate another approach relying on an ontological representation learning strategy based on definitions, and the empirical properties of semantic latent spaces, described by \citet{nandakumar-etal-2019-well} and \citet{garcia-etal-2021-probing}. In particular, we investigate whether semantic models trained from ontological definitions perform better than other semantic models for the task of identifying idiomatic MWEs without relying on their usage in context, using a novel self-explainability score which will be introduced in Section \ref{sec:methodology}. 

\section{Methodology}
\label{sec:methodology}

After collecting multiword entity names, a chosen semantic model is used to map the obtained MWEs $(W_1...W_n)$ to their latent 
representations, either as a whole $(\RS)$ or word-per-word $(\R{i})$.
\begin{equation*}
\begin{split}
    & \RS   := \textrm{SemReprOf}(W_1...W_n) \\
    & \R{i}\hspace{0.1cm} := \textrm{SemReprOf}(W_i) \\
\end{split}
\end{equation*}

Our semantic model, being based on a Transformer + Mean Pooling pipeline (see Figure \ref{fig:model}), produces its representations by averaging the representation of the tokens it is provided as an input (after taking their interactions into account):
\begin{equation*}
\begin{split}
&\RS = \frac{1}{n}\sum{\textrm{SemReprOf}(W_i|W_1...W_n)}
\end{split}
\end{equation*}

To isolate the effect of these interactions, we compute a weighted average of the independent representations of the constituents of the MWE (with weights $\alpha_i$) as a generalization of the above:
\begin{equation*}
\begin{split}
{\textstyle\sum\alpha_i\overline{\textsc{r}_i}} = \sum{\alpha_i\textrm{SemReprOf}(W_i)}
\end{split}
\end{equation*}

Our novel self-explainability score for MWEs corresponds to the degree of similarity between their latent semantic representation ($\overline{\textsc{r}_\Sigma}$) and the best\footnote{We determine the optimal weights $\alpha_i$ in Appendix \ref{sec:appendix_weights}.} weighted average of the independent representations of their constituents $(\sum\alpha_i\overline{\textsc{r}_i})$. 
\begin{equation*}
\begin{split}
    & \mathcal{S} := \max_{\alpha_i}\big[\mathrm{CosSim}(\textstyle \sum\alpha_i\overline{\textsc{r}_i}, \RS)\big]
\end{split}
\end{equation*}


Only strong inter-constituent interactions should be able to explain low self-explainability scores. 

Based on this insight, we hypothesize that low self-explainability scores identify the MWEs that the semantic model treats as idiomatic. To validate our hypothesis, we will demonstrate that there is indeed a statistically significant difference in self-explainability scores between idiomatic and non-idiomatic MWEs, among a chosen population.

For our analysis, we construct a set of two-words MWEs obtained from UMLS\footnote{All two-words entity names from UMLS were included, after filtering out pairs containing words which are either too frequent (>10000 occurences) or too rare (<10 occurences) in the UMLS ontology. This amounts to about 100 thousand two-words MWEs (98.307 to be precise).}, which were then subsequently divided into two groups by our annotators\footnote{The labelling was performed by two annotators: a trained linguist specialized in MWEs who is currently following a course on medical translation, and a NLP practitioner with multiple years of experience in clinical NLP (with an inter-annotator agreement of 82.5\% and a kappa score of 0.54).}: those which were “perceived as idiomatic or semi-idiomatic” and those which were “perceived as self-explanatory”.

We also hypothesize that a definition-based pre-training is essential for this analysis to produce good results. However, as the proposed analysis could be applied to any contextual text representation model,
we set out to evaluate the benefits of the definition-based pretraining of the BioLORD model \citep{remy-etal-2022-biolord} by comparing its results with two strong alternatives: SapBERT \citep{liu-etal-2021-self} and CODER \citep{Yuan2022}. These two state-of-the-art biomedical language models were also trained using contrastive learning and UMLS, but not using definitions as a semantic anchor.

\begin{figure}
    \centering
    \includegraphics[width=0.41\textwidth]{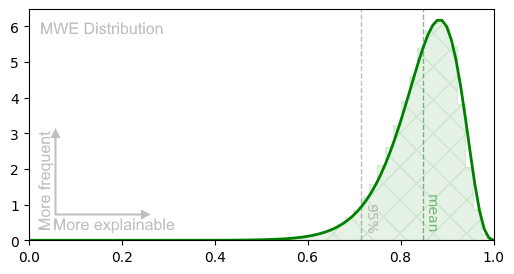}
    \caption{Density of self-explainability scores produced by BioLORD for all the MWEs of our dataset.}
    \label{fig:full_distribution}
    \vspace{0.345cm}
%
    \centering
    \includegraphics[width=0.41\textwidth]{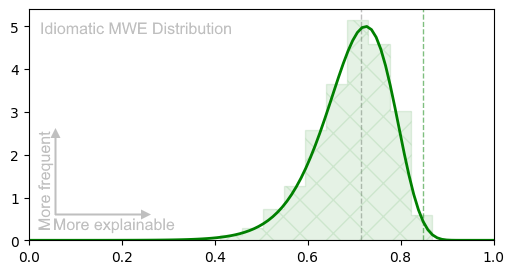}
    \caption{Density of self-explainability scores produced by BioLORD for the idiomatic MWEs.}
    \label{fig:idio_distribution}
    \vspace{0.345cm}
%
    \centering
    \includegraphics[width=0.41\textwidth]{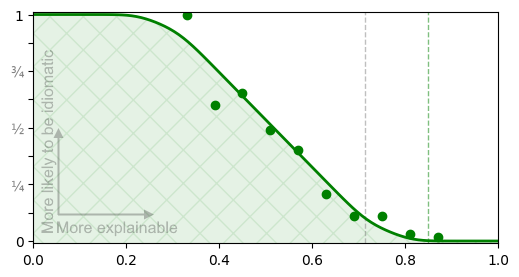}
    \caption{Proportion of MWEs preceived as idiomatic, in function of the self-explainability score produced by BioLORD (bullets represent our annotations).}
    \label{fig:idio_ratio}
    \vspace{0.345cm}
%
    \centering
    \includegraphics[width=0.41\textwidth]{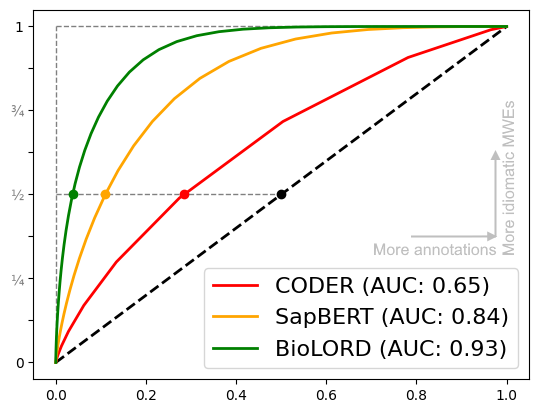}
    \caption{Comparison between the ROC curves of various biomedical models, which shows that BioLORD has a much large area under curve than the other models. The green dot represents the 95th percentile operating point described in the paper; this is the point where about half of the idiomatic MWEs are recalled; achieving the same result with the other models (orange and red dots) or through chance (black dot) requires processing multiple times more MWEs than BioLORD.\vspace{-4pt}}
    \label{fig:roc_curve}
\end{figure}


\section{Experimental Results}

We start our analysis by plotting the empirical distribution of self-explainability scores for all considered UMLS entities. We report this empirical distribution as a histogram in Figure \ref{fig:full_distribution}. 

Interestingly, this distribution is unimodal, which seems to give weight to the hypothesis that MWEs exist on a spectrum of idiomaticity, as described by \citet{Cowie1981}, and do not form clearly distinct idiomaticity classes. 

Based on our annotations, we evaluate the proportion of idiomatic MWEs present in a subset of 10 bins of self-explainability scores (see Figure \ref{fig:idio_ratio}). 

This enables us to estimate the full distribution of idiomatic MWEs by multiplying these ratios with the population counts (see Figure \ref{fig:idio_distribution}).

These two distributions have very different means (0.850 vs 0.697), indicating that our self-explainability score is indeed significantly lower for idiomatic MWEs than for non-idiomatic ones.



We determined based on our annotations that about 2.6\% of the MWEs in our dataset appeared idiomatic or semi-idiomatic in nature. To evaluate how effectively our self-explainability score can help identifying idiomatic MWEs, we determined the threshold score enabling a recall of about 50\% of idiomatic MWEs in our dataset. This corresponds to about 4000 MWEs featuring a similarity below 0.714, consisting of the outliers at or below the 95\% percentile of our self-explainability scores.

To confirm this, we annotated more extensively the MWEs of our dataset falling into these 5 outlier percentiles. We find that about 23\% of these MWEs appear idiomatic to our annotators, which is in line with our population-based estimates of 26\% (2.6\% of idiomatic MWEs * 50\% recall = 1.3\% of idiomatic MWEs out of these 5\% of outliers, yielding an expected precision of 26\%). 

Of course, a threshold of 0.714 represents only one of the possible operating points of our model. By varying this threshold, we compute the receiver operating characteristic (ROC) of our classifier, and plot it in Figure \ref{fig:roc_curve} (green curve). We find that our model shows an area under curve (AUC) of 93\%. 

\begin{figure}[b!]
    \centering
    \includegraphics[width=0.41\textwidth]{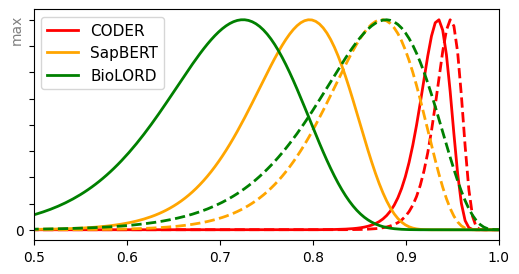} 
    \caption{Density of self-explainability scores produced by the compared models for the idiomatic (solid) and self-explainable (dotted) MWEs of our dataset.}
    \label{fig:sapbert_distrib}
    \vspace{0.35cm}
\end{figure}

Repeating this analysis for other semantic biomedical models demonstrates the importance of BioLORD's definition-based training. Indeed, both SapBERT (orange curve) and CODER (red curve) fail to provide a classifier that is as effective as BioLORD for this task, with AUC scores of 0.84 and 0.65 respectively. See also Figure \ref{fig:sapbert_distrib}.


To enable a more qualitative appreciation of the results, we also report the MWEs featuring the lowest self-explainability scores, for each of the considered models (see Table \ref{tab:summary}). 
Based on this, we note that the outliers of the BioLORD model are not only of higher quality, but also feature a significantly lower self-explainability scores. We interpret this as an indication that, to produce definition-grounded representations for MWEs, the BioLORD model has to devote more of its weights to memorize and specialize idiomatic MWEs than the other models.

We can further this impression by looking at Figure \ref{fig:sapbert_distrib}. While SapBERT has a distribution of scores similar to BioLORD, the difference between the idiomatic and self-explanatory MWEs is less pronounced, leading to more mixups. Looking further, we also notice that the CODER model seems to feature almost no score variation between MWEs in general, and appears to treat few MWEs as idiomatic (besides a few general-purpose hold-outs from its original pre-training). These findings again comfort the idea that a definition-based pre-training is important to achieve good results.

\begin{table}[b!]
    \centering
    \begin{tabular}{|l|lr|}
        \hline
        \textbf{Model} & \textbf{MWE} & $\mathcal{S}$\textbf{-score} \\
        \hline
        \multirow{3}{*}{{BioLORD}} & Gray Matter & 0.30 \\
        & Neprogenic rest & 0.32 \\
        & Heyman operation & 0.33 \\
        \hline
        \multirow{3}{*}{{SapBERT}} & Ibuprofen dose & 0.49 \\
        & Anal Lymphoma & 0.53 \\
        & Hemoglobin Wood & 0.54 \\
        \hline
        \multirow{3}{*}{{CODER}} & United Kingdom & 0.75 \\
        & Small Molecule & 0.77 \\
        & United States & 0.78 \\
        \hline
    \end{tabular}
    \caption{Most extreme self-explainability outliers for the models compared in this study. An extended version of this table can be found in Appendix \ref{sec:appendix_examples}.}
    \label{tab:summary}
\end{table}

\newpage
\section{Conclusion}
In this paper, we investigated the suitability of definition-based semantic models for detecting idiomatic MWEs in the terminology of a domain. We were able to demonstrate that our proposed self-explainability score can indeed serves as a proxy for idiomaticity, and observed that the BioLORD model indeed displays strong ability to perform this evaluation in the biomedical domain. 

The corpus-free idiomaticity estimation thereby developed is powerful enough to help ontology translators to focus on more challenging MWEs, with about half of the idiomatic MWEs contained in the 5\% of self-explainability score outliers.

Finally, we were also able to show that biomedical models which were not trained using a definition-based strategy perform significantly worse than our chosen definition-based model, showing the importance of a definition-based pre-training strategy in the development of reliable semantic representations for idiomatic MWEs.


\section*{Limitations}

It is worth noting that the approach described in this paper can only be expected to operate reliably on entities which can be accurately represented in the latent space by the chosen semantic model (either through its exposure to textual definitions or ontological relationships about the entity during pre-training, or through its generalization abilities). 

Unlike past approaches for detecting idiomatic MWEs, our strategy cannot make use of context to recognize idiomatic MWEs from their usage in a corpus. It would be an interesting future work to investigate how to combine examples of uses and ontological knowledge to develop a better in-context idiomaticity evaluation for MWEs.

An additional limitation of our work, is that we limited our analysis to UMLS entities consisting of exactly two words. This is not a limitation of our proposed approach per se, but we acknowledge that further work should probably be carried out to investigate how to best handle longer sequences.

\section*{Ethics Statement}

The authors of this paper do not report any particular ethical concern regarding its content.

%

\newpage
\bibliography{anthology,custom}
\bibliographystyle{acl_natbib}

\clearpage
\appendix
\renewcommand\thefigure{\thesection.\arabic{figure}}
\renewcommand{\thetable}{\thesection.\arabic{table}}
\setcounter{table}{0}
\setcounter{figure}{0}    

\section{An analytical solution for the optimal vector averaging problem}
\label{sec:appendix_weights}

In this appendix, we derive the analytical solution for the problem of finding the optimal weighted average (of the representation of the constituents of a MWE) given the task of maximizing the cosine similarity between their weighted average and the representation of the MWE itself.

Let $\R{1}$ and $\R{2}$ be two vectors (the representation of the words $W_1$ and $W_2$ through the BioLORD model).
Let $\RS$ be a vector (the representation of the MWE through the BioLORD model).

\begin{figure}[t]
    \centering
    \includegraphics[width=0.408\textwidth]{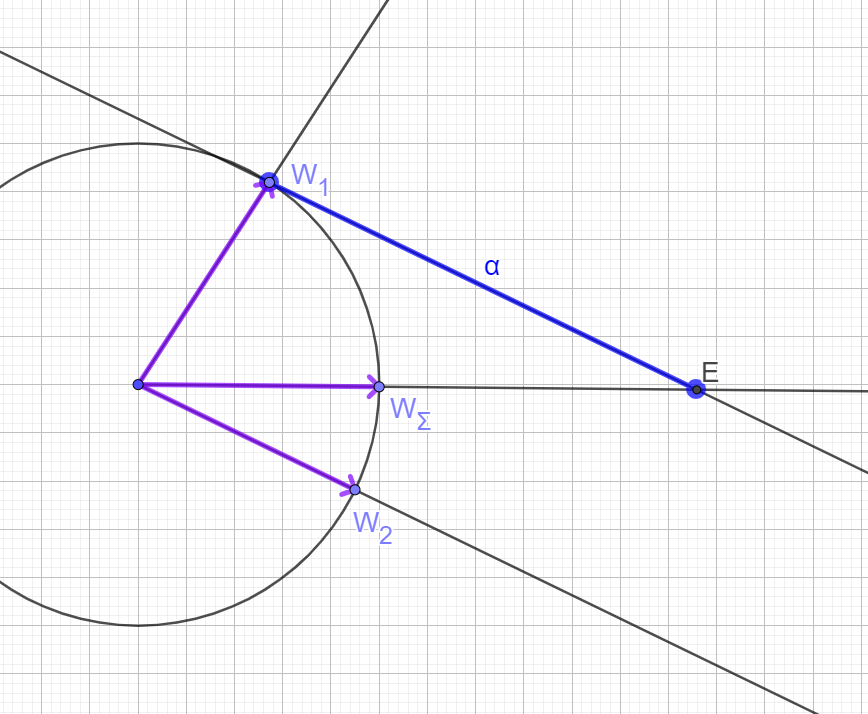}
    \caption{Representation of the problem}
    \label{fig:alpha}
\end{figure}

\begin{center}
\noindent{\color{LightGray}\small ... see Figure \ref{fig:alpha} ...}
\end{center}

Our objective is to maximize the cosine similarity between $\RS$ and a weighted average of the vectors $\R{i}$ (with weights ${\alpha}_i$).
Because the cosine similarity between two vectors does not depend on their respective lengths, we can without loss of generality try to maximize the following expression for the mixing parameter ${\alpha}={\alpha}_2/{\alpha}_1$.
\begin{equation*}
    \textit{CosSim}(\R{1}+\alpha\R{2},\RS ) := \frac{(\R{1}+\alpha\R{2})\cdot(\RS)}{|\R{1}+\alpha\R{2} |.|\RS |}
\end{equation*}

Because the maximum cosine similarity will necessarily be positive, we can look for the maximum of its square instead. We will find our optimum by looking at the points where the derivative is equal to 0:
\begin{equation*}
    \frac{d}{d\alpha}\big[\textit{CosSim}^2(\R{1}+\alpha\R{2},\RS)\big]=0
\end{equation*}

\begin{center}
    
\noindent{\color{LightGray}\small ... recalling $\frac{d}{dx}\big[\frac{f}{g}\big]=\big[g\frac{df}{dx} - f \frac{dg}{dx}\big] / \big[g^2\big]$ ...}
\begin{equation*}  
\begin{gathered}
    (\R{1}+\alpha\R{2} )^2  \frac{d}{d\alpha}\big[((\R{1}+\alpha\R{2} )\cdot(\RS ))^2\big] \\
    = ((\R{1}+\alpha\R{2} )\cdot(\RS ))^2  \frac{d}{d\alpha}\big[(\R{1}+\alpha\R{2} )^2\big]
\end{gathered}
\end{equation*}

\noindent{\color{LightGray}\small ... computing the inner derivatives ...}
\begin{equation*}
\begin{gathered}
(\R{1}+\alpha\R{2} )^2 (2((\R{1}+\alpha\R{2} )\cdot(\RS ))(\R{2}\cdot\RS )) \\
= ((\R{1}+\alpha\R{2} )\cdot(\RS ))^2 (2(\R{1}+\alpha\R{2} )(\R{2}))    
\end{gathered}
\end{equation*}

\noindent{\color{LightGray}\small ... dividing both sides by $2$ and $(\R{1}+\alpha\R{2})(\RS)$ ...}
\begin{equation*}
\begin{gathered}
    (\R{1}+\alpha\R{2} )^2 (\R{2}\cdot\RS ) \\
    = ((\R{1}+\alpha\R{2} )\cdot(\RS ))((\R{1}+\alpha\R{2} )\cdot(\R{2} ))    
\end{gathered}
\end{equation*}

\end{center}

\newpage

Let’s introduce a more convenient notation for the scalar products ($\textsc{r}_{xy}=\R{x}\cdot\R{y}$). Given we are trying to find scaling coefficients for $\R{i}$ vectors, we can first normalize them to make their norm is equal to one, without loss of generality, such that $\textsc{r}_{11}=\textsc{r}_{22}=\textsc{r}_{\Sigma\Sigma}=1$. 

\begin{center}

\noindent{\color{LightGray}\small ... expanding the products ...}
\begin{equation*}
\begin{gathered}
(\textsc{r}_{11} + 2\alpha \textsc{r}_{12} + \alpha^2 \textsc{r}_{22})(\textsc{r}_{2\Sigma}) \\
= (\textsc{r}_{1\Sigma} \textsc{r}_{12} + \alpha \textsc{r}_{2\Sigma} \textsc{r}_{12} + \alpha \textsc{r}_{1\Sigma} \textsc{r}_{22} + \alpha^2  \textsc{r}_{2\Sigma} \textsc{r}_{22})        
\end{gathered}
\end{equation*}

\noindent{\color{LightGray}\small ... isolating $\alpha$ on the left side ...}
\begin{equation*}
\alpha(\textsc{r}_{12} \textsc{r}_{2\Sigma} - \textsc{r}_{1\Sigma} \textsc{r}_{22}) = (\textsc{r}_{1\Sigma} \textsc{r}_{12} - \textsc{r}_{11} \textsc{r}_{2\Sigma})    
\end{equation*}

\noindent{\color{LightGray}\small ... giving us the formula of $\alpha$ ...}
\begin{equation*}
    {\alpha}=\frac{\textsc{r}_{1\Sigma} \textsc{r}_{12} - \textsc{r}_{2\Sigma} \textsc{r}_{11}}{\textsc{r}_{2\Sigma} \textsc{r}_{12} - \textsc{r}_{1\Sigma}  \textsc{r}_{22}} = \frac{\textsc{r}_{1\Sigma} \textsc{r}_{12} - \textsc{r}_{2\Sigma}}{\textsc{r}_{2\Sigma}  \textsc{r}_{12} - \textsc{r}_{1\Sigma}}
\end{equation*}

\noindent{\color{LightGray}\small ... giving us the formula of $\alpha_i>0$ ...}
\begin{equation*}
\begin{gathered}
    {\alpha}_1 = \textsc{r}_{1\Sigma} - \textsc{r}_{12} \textsc{r}_{2\Sigma} \\
    {\alpha}_2 = \textsc{r}_{2\Sigma} - \textsc{r}_{21} \textsc{r}_{1\Sigma}
\end{gathered}
\end{equation*}
\end{center}

\textbf{Intuition:} If we assume that the constituents of the entity have orthogonal meanings ($\R{1}\cdot\R{2}=0$), this gives $\alpha_1=\textsc{r}_{1\Sigma}$ and $\alpha_2=\textsc{r}_{2\Sigma}$ which are the cosine similarities of each constituent with respect to the entire MWE.

\newpage
\twocolumn[\section{Examples of similarity outliers for the considered models}]
\label{sec:appendix_examples}

\begin{table}[t!]
\centering
\begin{tabular}{lll}
\textbf{Word1} & \textbf{Word2} & \textbf{Score} \\
Gray           & Matter         & 0.303302       \\
Nephrogenic    & rest           & 0.317366       \\
Heyman         & operation      & 0.328952       \\
Chemical       & procedure      & 0.331814       \\
Morning        & sickness       & 0.359685       \\
Morning        & Sickness       & 0.359685       \\
Green          & Card           & 0.364002       \\
Yellow         & Fever          & 0.365865       \\
Nitrogen       & retention      & 0.372655       \\
molecular      & function       & 0.374572       \\
osseous        & survey         & 0.384946       \\
Refsum         & Disease        & 0.38831        \\
Monteggia's    & Fracture       & 0.392137       \\
Silver         & operation      & 0.393802       \\
Worth          & disease        & 0.395263       \\
Diseases       & Component      & 0.398678       \\
Root           & stunting       & 0.402461       \\
McBride        & operation      & 0.403504       \\
Air            & hunger         & 0.405719       \\
Storage        & disease        & 0.414184       \\
Border         & Disease        & 0.415117       \\
Intersection   & syndrome       & 0.417804       \\
Retinal        & correspondence & 0.420826       \\
Patch          & Testing        & 0.423289       \\
Dot            & haemorrhages   & 0.423748       \\
Coordination   & Complexes      & 0.4248         \\
White          & matter         & 0.426788       \\
Molar          & concentration  & 0.432153       \\
Book           & Syndrome       & 0.432465       \\
Circulatory    & depression     & 0.4349         \\
German         & Syndrome       & 0.436444       \\
Nissen         & Operation      & 0.438874       \\
Physical       & shape          & 0.440117       \\
External       & features       & 0.442601       \\
Anoxic         & neuropathy     & 0.443183       \\
Compartment    & syndromes      & 0.445978       \\
Visceral       & Myopathy       & 0.447205       \\
Tumour         & haemorrhage    & 0.447391       \\
Mountain       & Sickness       & 0.44767        \\
Growth         & Factor         & 0.451592      
\end{tabular}
\caption{Self-explainability outliers for BioLORD}
\label{tab:outliers_biolord}
\end{table}

\begin{table}[t!]
\centering
\begin{tabular}{lll}
\textbf{Word1}  & \textbf{Word2} & \textbf{Score} \\
ibuprofen       & dose           & 0.488790       \\
Anal            & Lymphoma       & 0.531192       \\
Hemoglobin      & Wood           & 0.542635       \\
Ovarian         & injury         & 0.548922       \\
Ovarian         & perforation    & 0.557121       \\
Ibuprofen       & overdose       & 0.569412       \\
hemoglobin      & Aurora         & 0.575010       \\
miconazole      & injection      & 0.575241       \\
diphenhydramine & Cartridge      & 0.580044       \\
phenylephrine   & Injection      & 0.584401       \\
Hemoglobin      & Mexico         & 0.585959       \\
Dexamethasone   & Powder         & 0.589987       \\
Hydrocortisone  & phosphate      & 0.592702       \\
Guaifenesin     & poisoning      & 0.592808       \\
hydrocortisone  & receptor       & 0.594878       \\
Vaginal         & adenocarcinoma & 0.595991       \\
iv              & lidocaine      & 0.598489       \\
Gonadal         & Thrombosis     & 0.598919       \\
Rectal          & artery         & 0.603538       \\
hemoglobin      & Cook           & 0.606404       \\
Ibuprofen       & Powder         & 0.606984       \\
hemoglobin      & Thailand       & 0.608336       \\
Ovarian         & vessels        & 0.609299       \\
Intestinal      & hematoma       & 0.610457       \\
diphenhydramine & Injection      & 0.611432       \\
hemoglobin      & Chicago        & 0.611646       \\
Ornithine       & Ql             & 0.612263       \\
Aspirin         & dose           & 0.613269       \\
Hydrocortisone  & Injection      & 0.613701       \\
Ovarian         & hematoma       & 0.613911       \\
hemoglobin      & Oita           & 0.614288       \\
Wrist           & injection      & 0.614621       \\
Hemoglobin      & Ohio           & 0.614865       \\
Aspirin         & overdose       & 0.615012       \\
Oral            & hemangioma     & 0.615188       \\
Hemoglobin      & Shanghai       & 0.618727       \\
Sodium          & retention      & 0.619068       \\
Diphenhydramine & overdose       & 0.619255       \\
hemoglobin      & Bristol        & 0.619368       \\
Gonadal         & artery         & 0.620956      
\end{tabular}
\caption{Self-explainability outliers for SapBERT\vspace*{1.31in}}
\label{tab:outliers_sapbert}
\end{table}

\begin{table}[t!]
\centering
\begin{tabular}{lll}
\textbf{Word1} & \textbf{Word2} & \textbf{Score} \\
United         & Kingdom        & 0.754104       \\
Small          & Molecule       & 0.772967       \\
United         & States         & 0.775555       \\
Dependent      & Variable       & 0.796870       \\
patch          & clamp          & 0.799848       \\
Index          & finger         & 0.809509       \\
Eggshell       & nail           & 0.810650       \\
single         & molecule       & 0.812445       \\
Data           & Administration & 0.818826       \\
Alkaline       & Phosphatase    & 0.818921       \\
Brush          & Border         & 0.820135       \\
Czech          & Republic       & 0.821894       \\
CrAsH          & compound       & 0.822972       \\
Nuclear        & medicine       & 0.823420       \\
Nuclear        & Medicine       & 0.823420       \\
Hydrogen       & Bonds          & 0.823888       \\
Replication    & Origin         & 0.825065       \\
Wild           & Type           & 0.825602       \\
Antigen        & Presentation   & 0.826336       \\
outer          & membrane       & 0.827730       \\
Inclusion      & Bodies         & 0.829212       \\
Health         & administration & 0.829440       \\
Active         & Site           & 0.829467       \\
Focus          & Groups         & 0.830125       \\
Natural        & killer         & 0.830615       \\
Click          & Chemistry      & 0.831714       \\
Strand         & breaks         & 0.832437       \\
proc           & gene           & 0.832669       \\
Lewis          & antigen        & 0.833199       \\
lucifer        & yellow         & 0.833356       \\
Mass           & Spectrometry   & 0.833356       \\
Foreign        & Bodies         & 0.833412       \\
Foreign        & body           & 0.833504       \\
Uvea           & language       & 0.836055       \\
Williams       & Syndrome       & 0.836802       \\
pyridoxine     & clofibrate     & 0.837463       \\
Precision      & Medicine       & 0.838389       \\
Antigen        & Switching      & 0.838619       \\
Public         & Domain         & 0.838712       \\
Data           & Acquisition    & 0.838931
\end{tabular}
\caption{Self-explainability outliers for CODER\vspace*{1.7075in}}
\label{tab:outliers_coder}
\end{table}


\begin{figure}[b!]
    \section*{Acknowledgements}
    
This work would not have been possible without the joint financial support of the Vlaams Agentschap Innoveren \& Ondernemen (VLAIO) and the RADar innovation center of the AZ Delta hospital group. 

\vspace{0.3cm}

We also want to thank the SIGLEX-MWE 2023 (19th Workshop on Multiword Expressions) and the ClinicalNLP 2023 Workshop for organizing the joint session to which this paper was submitted, which enabled the blooming of this collaboration between machine learning engineers and linguists
thanks to its existence.

    \vspace{25cm}
    
\end{figure}

\end{document}